\documentclass[sigconf]{acmart}
\renewcommand\footnotetextcopyrightpermission[1]{} % 
\usepackage{listings}
\usepackage{algorithm}
\usepackage{algpseudocode}
\usepackage{arydshln}
\usepackage{amsmath}
\usepackage{listings}
\usepackage{xcolor}
\usepackage{tcolorbox}
\usepackage{fancyvrb}
\usepackage{mdframed}
\usepackage{booktabs}
\usepackage{multirow}
\usepackage{xcolor}
\usepackage{colortbl}
\usepackage{array}
\usepackage{siunitx}
\usepackage{graphicx}
\usepackage{adjustbox}
\AtBeginDocument{%
  }

\begin{document}
\author{Minh-Anh Nguyen}
\authornote{Both authors contributed equally to this research.}
\affiliation{%
  \institution{CMC OpenAI, VinUniversity
  }
  \country{Vietnam}
}

\author{Minh-Duc Nguyen}
\authornotemark[1]
\affiliation{%
  \institution{VinUniversity}
  \country{Vietnam}
}

\author{Ha Lan N.T.}
\authornotemark[1]
\affiliation{%
  \institution{VinUniversity}
  \country{Vietnam}}

\author{Kieu Hai Dang}
\affiliation{%
  \institution{VinUniversity}
  \country{Vietnam}
}
\author{Nguyen Tien Dong}
\affiliation{%
  \institution{CMC OpenAI, VinUniversity
  }
  \country{Vietnam}
}
\author{Dung D. Le}
\affiliation{%
  \institution{VinUniversity}
  \country{Vietnam}
}
\renewcommand{\shortauthors}{Minh-Anh et al.}
\title{A Multi-Agent LLM Framework with Hierarchical Citation Graph for Automated Survey Generation}
\begin{abstract}
Large language models (LLMs) are increasingly being adopted to automate survey paper generation. Existing approaches typically extract content from a large collection of related papers and prompt LLMs to summarize them directly. However, such methods often overlook the structural relationships among papers, resulting in generated surveys that lack a coherent taxonomy and a deeper contextual understanding of research progress. To address these shortcomings, we propose \textbf{SurveyG}, an LLM-based agent framework that integrates \textit{hierarchical citation graph}, where nodes denote research papers and edges capture both citation dependencies and semantic relatedness between their contents, thereby embedding structural and contextual knowledge into the survey generation process. The graph is organized into three layers: \textbf{Foundation}, \textbf{Development}, and \textbf{Frontier}, to capture the evolution of research from seminal works to incremental advances and emerging directions. By combining horizontal search within layers and vertical depth traversal across layers, the agent produces multi-level summaries, which are consolidated into a structured survey outline. A multi-agent validation stage then ensures consistency, coverage, and factual accuracy in generating the final survey. Experiments, including evaluations by human experts and LLM-as-a-judge, demonstrate that SurveyG outperforms state-of-the-art frameworks, producing surveys that are more comprehensive and better structured to the underlying knowledge taxonomy of a field. Our implementation can be found at \url{https://github.com/akBear23/SurveyG.git}.
\end{abstract}

\begin{CCSXML}
<ccs2012>
   <concept>
       <concept_id>10010147.10010178.10010179.10003352</concept_id>
       <concept_desc>Computing methodologies~Information extraction</concept_desc>
       <concept_significance>500</concept_significance>
       </concept>
 </ccs2012>
\end{CCSXML}
\ccsdesc[500]{Computing methodologies~Information extraction}
\keywords{Automated Survey Generation, Literature Synthesis}

%%
%% This command processes the author and affiliation, and title
\maketitle

\begin{figure}[h]
  \centering
  \includegraphics[width=\columnwidth]{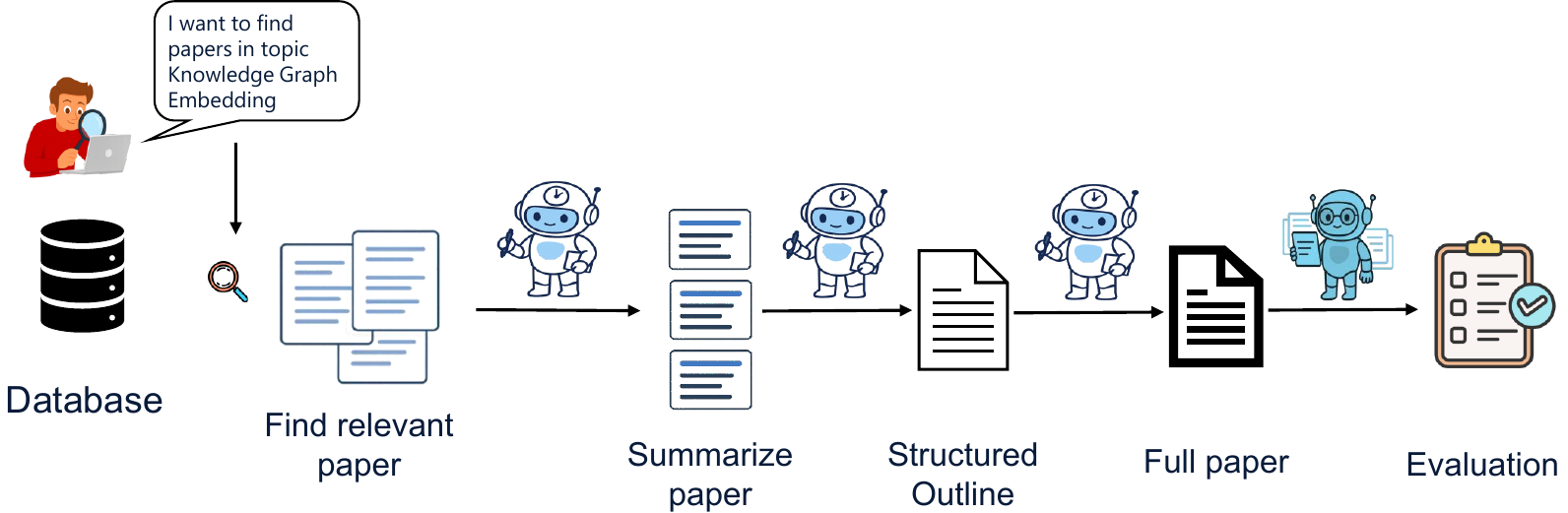}
\caption{Standard automated survey pipeline: (1) collect relevant papers, (2) build a structured outline, and (3) write the full survey.}
  \label{fig:basic pipeline}
\end{figure}

\section{Introduction}
The exponential growth of research publications, particularly in rapidly evolving fields such as Artificial Intelligence \cite{huynh2023artificial}, has made it increasingly difficult for researchers to keep pace with new developments \cite{wang2406autosurvey, wen2025interactivesurvey}. While survey papers serve as invaluable resources by synthesizing existing knowledge and identifying emerging trends, their manual construction is costly, time-consuming, and often unable to keep up with the overwhelming influx of literature \cite{liang2025surveyx}. Although large language models (LLMs) offer promising text generation capabilities, they face critical limitations in handling massive reference sets, maintaining academic rigor, and providing up-to-date knowledge \cite{wu2025inference, han2024llm}. These challenges underscore the urgent need for an automated survey generation framework that can efficiently retrieve, organize, and synthesize literature into coherent, high-quality surveys tailored to users’ research interests.

Some recent studies \cite{su2025benchmarking, wang2406autosurvey, liang2025surveyx, yan2025surveyforge} have proposed autonomous survey generation frameworks based on user queries, following the basic pipeline illustrated in Figure \ref{fig:basic pipeline}. While these approaches represent promising progress, they exhibit two key limitations. \textbf{Firstly,} they neglect the relationships between papers, such as citation links, methodological connections, or subtopic dependencies, which are essential for understanding how works build upon one another, improve over foundational methods, and collectively shape research trends. \textbf{Secondly,} these frameworks employ a naive strategy for constructing structured outlines or full survey papers, simply concatenating summaries of individual papers. This not only exacerbates the long-context problem in LLMs but also fails to exploit the hierarchical organization of related works within subtopics.

To address these limitations, we propose \textbf{SurveyG}, an autonomous survey generation system that emphasizes knowledge representation and hierarchical summarization. We design an LLM-based multi-agent framework that represents knowledge using a hierarchical citation graph, where nodes correspond to papers and edges capture citation relationships and semantic similarity. The graph is organized into three layers: \textbf{Foundation, Development}, and \textbf{Frontier}, reflecting research progression from seminal contributions to emerging directions. By combining horizontal searches within layers and vertical traversals across layers, our framework generates multi-aspect summaries that are consolidated into structured survey outlines. We evaluate SurveyG on 10 computer science topics from the SurGE benchmark \cite{su2025benchmarking}, comparing against state-of-the-art frameworks \cite{liang2025surveyx,yan2025surveyforge,wang2406autosurvey} along five dimensions: \textit{Coverage}, \textit{Structure}, \textit{Relevance}, \textit{Synthesis}, and \textit{Critical Analysis}. Our key contributions are: (1) a hierarchical citation graph representation modeling both citation and semantic relationships, (2) a graph-based traversal mechanism that produces diverse, multi-aspect summarizations capturing methodological foundations, developmental trends, and frontier directions, and (3) a multi-agent framework combining retrieval-augmented generation with hierarchical summaries as memory to construct coherent, evidence-grounded survey drafts.
\section{Methodology}
\begin{figure*}[h]
  \centering
  \includegraphics[width=\textwidth]{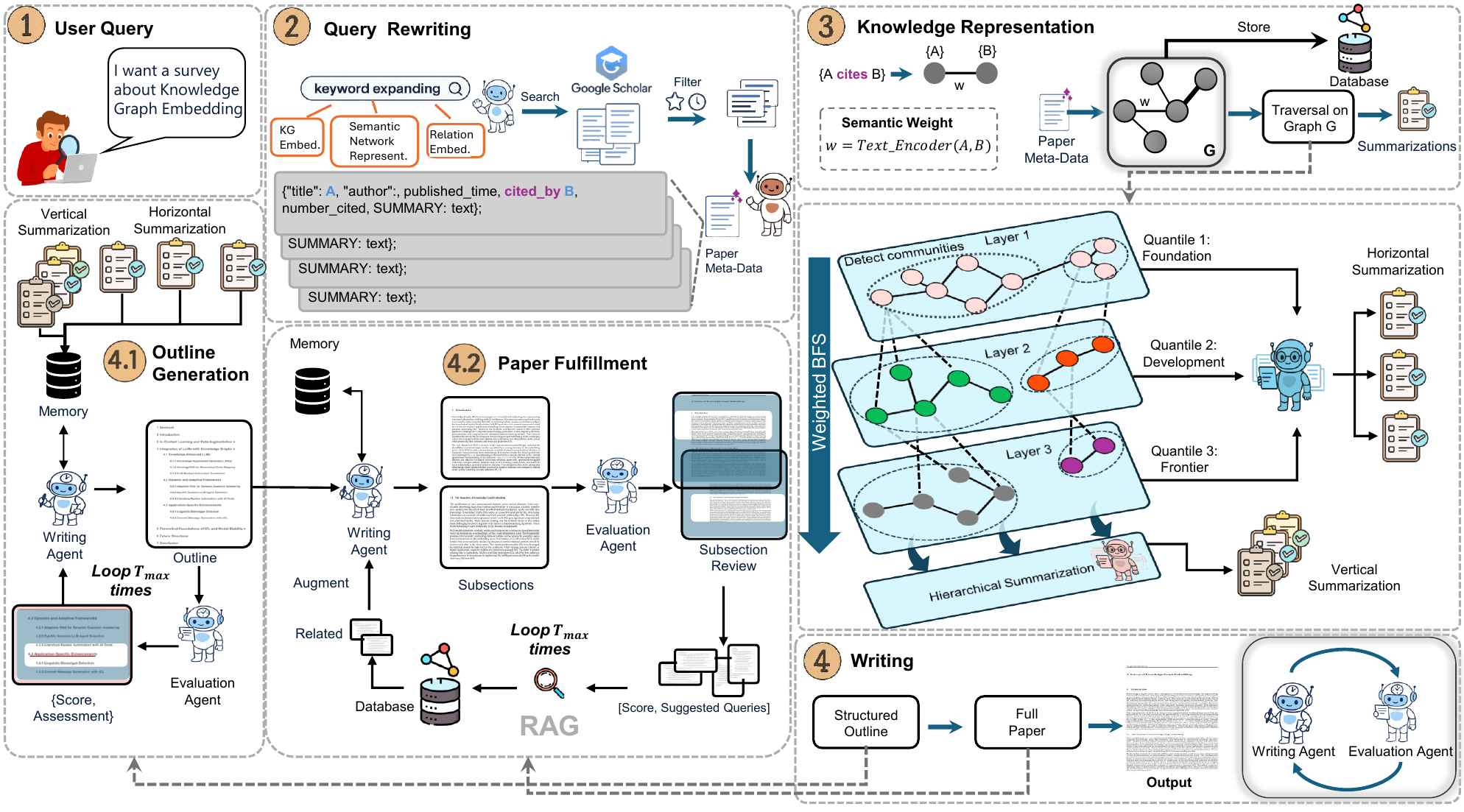}
  \caption{Starting from a user’s query, SurveyG retrieves and filters relevant papers (step 1-2), builds a hierarchical citation graph, and applies horizontal and vertical traversals to produce multi-aspect summaries (step 3). A multi-agent framework then leverages these pre-built summaries to produce a structured outline and a complete survey paper (step 4).}
  \label{overview pipeline}
\end{figure*}
We introduce \textbf{SurveyG}, an automated survey generation framework with two phases: a preparation phase that retrieves and summarizes papers, builds a hierarchical citation graph, and extracts relationships, and a generation phase that creates a structured outline and composes the full survey through instruction-prompted multi-agent collaboration (Figure~\ref{overview pipeline}).
\subsection{Preparation Phase}\label{prepare}
We represent the relationships among papers using a \textit{hierarchical citation graph}, where nodes correspond to academic papers and edges capture both citation links and semantic similarity, each weighted by a value $w$. Each node is further assigned to one of three layers: \textbf{Foundation}, \textbf{Development}, or \textbf{Frontier}, which reflect the role of the paper in the progression of research. Formally, the hierarchical citation graph is defined as $G = (V, E, L),$
where $V$ denotes the set of nodes (papers), $E \subseteq V \times V$ is the set of directed or undirected edges encoding citation or semantic relationships, and $$L: V \to \{\text{Foundation}, \text{Development}, \text{Frontier}\}$$ is a layer assignment function. For each node $v_i \in V$, we associate a corresponding document $d_i \in D$, where $D$ is a database storing the complete content of all papers. In addition, each node $v_i \in V$ is equipped with attributes 
that include a summarization of $d_i$ as well as metadata such as the paper’s title and publication year.
\subsubsection{Searching Relevant Paper}
Given a user query $q$, our goal is to construct a hierarchical citation graph $G$ that encompasses all relevant papers while capturing the evolutionary trends of research in the field. We first employ an LLM to expand the query into a set of diverse keywords $\{k_1, \ldots, k_n\} = \text{LLM}(q)$. Using these keywords, we retrieve candidate papers through the crawling module. After collecting the relevant papers, we establish edges between them based on citation links and quantify their semantic relatedness through weighted connections. The weight $w$ assigned to an edge connecting papers $v_i$ and $v_j$ is defined as  
\begin{equation}\label{edge_weight}
    w = \text{sim}\big(\text{Text\_Encoder}(v_i), \text{Text\_Encoder}(v_j)\big),
\end{equation}
where $\text{sim}(\cdot)$ denotes the cosine similarity between the text embedding vectors of the two papers. We consider two embedding strategies: abstract-based embeddings, which are efficient and widely available but may be incomplete or overstated, and LLM-generated summary-based embeddings, which better capture methodological and empirical content but require full text and extra processing \cite{zhang2025scientific}. We use summary-based embeddings by default because our pipeline already produces content-focused summaries (as detailed in the paragraph below), enabling richer semantic representations of methods, tasks, and domain contributions, and falls back to abstract-based embeddings when full-text access is unavailable.

To better leverage the key content of each paper for survey generation, every node $v_i$ is enriched with a summarization derived from its corresponding document $d_i$. We design specialized prompt templates tailored to different paper types \cite{liang2025surveyx} (e.g., surveys, methodological contributions, benchmarks, theoretical works) to extract salient information, including research objectives, methodological approaches, key findings, and limitations. After this phase, we obtain a flat graph $\hat{G} = (V, E)$ that encompasses the papers relevant to the user's topic along with their relationships. Each node $v_i \in V$ is associated with a set of attributes defined as
\[
A(v_i) = \{ \text{metadata}(v_i), \text{summary}(d_i) \},
\]
where $\text{metadata}(v_i)$ contains bibliographic information such as the paper's title, authors, and publication year, and $\text{summary}(d_i)$ represents the content-based summarization of the corresponding document. These attributes are also stored in the database $D$ to facilitate efficient retrieval and analysis.
\subsubsection{Knowledge Representation} 
To reflect the developmental progression of research within a topic, we assign each node in the flat graph $\hat{G}=(V,E)$ to one of three hierarchical layers via a layer assignment function. (1) \textbf{Foundation Layer}: The foundation layer consists of seminal and high-impact works that form the intellectual backbone of the field. For each paper $p$, we define a \textit{trending score} as
\begin{equation}\label{trending_score}
    \text{trend}_{\text{score}}(p) = \frac{\text{citation\_count}(p)}{1 + \text{year\_published}(p)},
\end{equation}
where $\text{year\_published}(p)$ denotes the number of years elapsed since the paper’s publication. Papers are ranked by this score, and the top-$K$ entries constitute the foundation set:
\[
    V_{\text{foundation}} = \{ v_i \in V \mid \text{trend}_{\text{score}}(v_i) \leq K \}.
\]
These papers are not only highly cited but also serve as conceptual anchors that establish key paradigms and problem formulations underpinning later research. (2) \textbf{Development Layer:} The development layer captures the historical evolution of the field before a time landmark $T$ (eg, 2025), representing works that refine, extend, or challenge the foundations. Formally, 
\[
    V_{\text{development}} = \{ v_i \in V \mid \text{year}(v_i) < T, \ v_i \notin V_{\text{foundation}} \},
\]  
These works are often incremental yet essential: they consolidate methodological frameworks, validate empirical findings, and enable the community to mature foundational ideas into established research threads. (3) \textbf{Frontier Layer:}  The frontier layer reflects the cutting edge of inquiry, consisting of recent contributions that point toward emerging trends and open challenges. It is defined as
    \[
    V_{\text{frontier}} = \{ v_i \in V \mid \text{year}(v_i) \geq T, \ v_i \notin V_{\text{foundation}} \}.
    \]  
Unlike the development layer, frontier works are temporally close to the present and thus provide a window into the current momentum and future trajectories of the domain. After this mapping, the hierarchical citation graph is represented as 
\[
G = (V, E, L), \quad V = V_{\text{foundation}} \cup V_{\text{development}} \cup V_{\text{frontier}}.
\]  
Traversing $G$ along horizontal (intra-layer) and vertical (inter-layer) edges then enables the generation of multi-aspect summaries covering methodologies, developmental trends, and future directions.
\begin{algorithm}[t]
\caption{Vertical Traversal for Multi-Summarization}
\label{alg:multi_traversal}
\begin{algorithmic}[1]
\State \textbf{Inp:} Citation graph $G=(V,E,L)$, foundation papers $V_{\text{foundation}}$
\State \textbf{Out:} $\{T_{\text{path}}^{(1)}, \dots, T_{\text{path}}^{(K)}\}$, where $K = |V_{\text{foundation}}|$
\ForAll{$s \in V_{\text{foundation}}$}
    \State $P_1 \gets \{\textsc{Extract}(s)\}$ 
    \State $P_2 \gets \{\textsc{Extract}(u) \mid u \in \textsc{WBFS}(s, \text{Development})\}$
    \State $P_3 \gets \{\textsc{Extract}(w) \mid w \in \textsc{WBFS}(P_2, \text{Frontier})\}$
    \State $T_{\text{dev}} \gets \textsc{GenerateSummarize}(P_1 \cup P_2)$
    \State $T_{\text{path}} \gets \textsc{GenerateSummarize}(T_{\text{dev}}, P_3)$
    \State Store $T_{\text{path}}$ as the summarization for seed $s$
\EndFor
\State \Return $K$ summarizations $\{T_{\text{path}}^{(1)}, \dots, T_{\text{path}}^{(K)}\}$
\end{algorithmic}
\end{algorithm}
\begin{figure}[t]
\begin{tcolorbox}[
    colback=gray!10, 
    colframe=black!40, 
]
You are a research analyst synthesizing papers on the topic \texttt{[QUERY]}.  
\texttt{<think>} Explain your reasoning for clustering papers into 2-3 subgroups based on methodology, contribution, or thematic focus. \texttt{</think>}  

For each subgroup, summarize the shared methodological approaches, thematic contributions, and provide a concise critique comparing the works.  
Finally, synthesize an overall perspective highlighting how these subgroups collectively operate in the field.
\end{tcolorbox}
\caption{Horizontal summarization short version prompt.}
\label{box:prompt}
\end{figure}
\subsubsection{Traversal on Graph Strategy}
We propose a two-stage summarization framework designed to capture both the breadth and depth of the hierarchical citation graph. In the \textit{horizontal stage}, to capture the internal structure of each layer $V_l$, we partition it into communities using the Leiden algorithm \cite{traag2019louvain}, yielding
\[
\mathcal{C}_l = \{ C_{l,1}, \ldots, C_{l,m_l} \}, \quad \bigcup_{j=1}^{m_l} C_{l,j} = V_l.
\]
Each community $C_{l,j}$ corresponds to a coherent research direction formed by citation and semantic proximity. For every community, we query an LLM using a carefully constructed prompt that integrates Plan-and-Solve~\cite{wang2023plan} strategies, along with paper-specific attributes, to generate a synthesized summary:
\[
T_{l,j} = \text{LLM}(\{ A(v_i) \mid v_i \in C_{l,j} \}),
\]
which emphasizes the methodologies and thematic scope of the papers while capturing key relationships among them. This process uncovers sub-directions within the topic and provides a global perspective of how research clusters evolve within each layer. The detailed prompts used for information extraction are provided in Figure~\ref{box:prompt}.

In the \textit{vertical stage}, we model cross-layer dependencies by tracing the evolution of ideas from each foundation paper. For every seed $s \in V_{\text{foundation}}$, Algorithm~\ref{alg:multi_traversal} performs a weighted breadth-first search (WBFS) over the citation graph, prioritizing semantically relevant nodes via edge weights. The procedure first extracts the foundation paper ($P_1$), then collects Development-layer papers reachable from $s$ ($P_2$), and finally expands to Frontier-layer papers from $P_2$ ($P_3$). We summarize these papers hierarchically: an intermediate summary $T_{\text{dev}}$ is generated from $P_1 \cup P_2$, which is then used to incorporate $P_3$ and produce the final path summary $T_{\text{path}}$. This staged design leverages temporal progression to reduce long-context issues and improve insight extraction \cite{zhang2024chain}. Repeating this process for all $K$ foundation papers yields $K$ vertical summaries, which are combined with $N$ horizontal layer summaries for a total of $K+N$ outputs.

% Add the differences
In contrast to earlier frameworks \cite{wang2406autosurvey, wen2025interactivesurvey, liang2025surveyx, yan2025surveyforge} that represent papers as isolated records in a flat database and depend exclusively on RAG-based retrieval, SurveyG organizes the literature within a hierarchical citation graph $G$. This representation integrates both citation and semantic connections among papers, allowing the system to capture the logical progression of research topics over time. By traversing this hierarchy, SurveyG produces a series of summarizations across multiple layers, effectively revealing methodological developments, evolutionary patterns, and current research frontiers. Such a design provides a more coherent and interpretable knowledge foundation for 
automated survey generation.
\begin{figure}[t]
\begin{tcolorbox}[
    colback=gray!10, 
    colframe=black!40, 
    % fontupper=\small
    % title=Development Path Summarization Prompt,
]
Create a comprehensive literature review outline based on the following taxonomy summaries for three layers (Foundation, Development, Frontier) and 
vertical directions.

\textbf{Horizontal summary:} \{summary\_layer\}.

\textbf{Vertical direction:} \{summary\_path\}. 

Based on the above information, create a detailed outline for a literature 
review paper, organizing it into sections and subsections.\textbf{ Respond with the outline in JSON format with keys: }

['section\_outline', 'subsection\_focus', 'proof\_ids'].

For each section: Put section and subsection titles in 'section\_outline'; Add a paragraph in 'subsection\_focus' describing the main focus of each subsection; Add 'proof\_ids' from either taxonomy layer or vertical direction.
\end{tcolorbox}
\caption{Structured outline creation short version prompt.}
\label{box:prompt_structured}
\end{figure}

\begin{algorithm}[t]
\caption{SurveyG Automated Survey Generation}
\label{alg:survey-generation}
\begin{algorithmic}[1]
\State \textbf{Input:} Survey Topic $Q$, Paper Database $D$, Max iterations $T_{\max}$, Summarizations $\{T_1, \ldots, T_{K+N}\}$
\State \textbf{Output:} Survey Paper $F$
\State \textit{// Initialization}
\State Create Writing Agent (WA) and Evaluation Agent (EA)
\State Initialize memory $M$ for WA with $\{T_1, \ldots, T_{K+N}\}$
\State \textit{// Phase 1: Create Outline}
\State WA generates initial outline $\mathcal{O}^{(0)}$ from $M$
\For{$t = 1$ to $T_{\max}$}
    \State $\mathcal{O}^{(t)} = \textsc{WA}(M, \textsc{EA}(\mathcal{O}^{(t-1)}))$
    \If{quality threshold met} \textbf{break}
    \EndIf
\EndFor
\State $\mathcal{O}^* \leftarrow \mathcal{O}^{(t)}$
\State \textit{// Phase 2: Write Full Paper}
\ForAll{subsection $O_{i} \in \mathcal{O}^*$}
    \State WA produces an initial draft $O_{i}^{(0)}$
    \State Refine with EA's feedback and suggested queries $Q$:
    \For{$t = 1$ to $T_{\max}$}
    \State \quad $O_{i}^{(t)} = \textsc{WA}(M \cup R_{i}^{(t)}, EA(O_{i}^{(t-1)}))$
    \State \quad where $R_{i}^{(t)} = \textsc{Retrieve}(Q_{i}^{(t)}, D)$
        \If{quality threshold met} \textbf{break}
        \EndIf
    \EndFor
\EndFor
\State \textit{// Phase 3: Assemble Survey}
\State $F \leftarrow \bigcup_{i} O_{i}^{(t)}$
\State \Return $F$
\end{algorithmic}
\end{algorithm}

\subsection{Generation Phase}\label{generation}
We use a multi-agent conversational framework \cite{wu2024autogen} for survey generation, consisting of a Writing Agent with memory initialized from $K+N$ graph-based summaries \cite{sumers2023cognitive} and an Evaluation Agent that provides diversity-focused feedback using LLM reasoning. Through iterative interaction, the Writing Agent drafts content grounded in summaries, while the Evaluation Agent critiques and refines it for coherence and balance, enabling collaborative survey construction, detailed in Algorithm~\ref{alg:survey-generation}.
\subsubsection{Structured Outline Construction}
The Writing Agent constructs an initial structured outline by grounding each section and subsection in the $K+N$ multi-aspect summarizations, ensuring both factual grounding and thematic coherence. The Evaluation Agent then reviews the draft, assessing logical flow and suggesting refinements without altering the overall structure. After the feedback iterations, the outline converges into a coherent and evidence-supported framework. Detailed prompting for both agents is provided in Figure \ref{box:prompt_structured} and Appendix~\ref{prompt}. The key innovation of SurveyG lies in its ability to manage long-context survey synthesis without concatenating all reference texts \cite{wang2406autosurvey, liang2025surveyx} or relying on pre-existing human-written surveys \cite{yan2025surveyforge}. Instead, it leverages hierarchical summarization from the citation graph $G$ as structured knowledge injected into the Writing Agent.
\begin{table*}[ht]
\centering
\small
\begin{tabular}{@{}llccccc@{}}
\toprule
\textbf{LLM} & \textbf{Model} & \textbf{Coverage} & \textbf{Structure} & \textbf{Relevance} & \textbf{Synthesis} & \textbf{Critical Analysis} \\
\midrule
\textbf{Claude}
& AutoSurvey & $73.6\,(\pm\,3.2)$ & $64.5\,(\pm\,4.1)$ & $80.2\,(\pm\,2.8)$ & $51.1\,(\pm\,5.3)$ & $45.8\,(\pm\,4.7)$ \\
& SurveyX & $74.2\,(\pm\,3.7)$ & $71.3\,(\pm\,4.3)$ & $82.3\,(\pm\,3.1)$ & $62.4\,(\pm\,4.9)$ & $52.7\,(\pm\,5.1)$ \\
& SurveyForge & $81.8\,(\pm\,2.9)$ & $78.4\,(\pm\,3.5)$ & $89.1\,(\pm\,2.1)$ & $75.4\,(\pm\,3.8)$ & $70.3\,(\pm\,4.2)$ \\
& SurveyG & $\textcolor{red}{\textbf{88.1}}\,(\pm\,2.4)$ & $\textcolor{red}{\textbf{87.9}}\,(\pm\,2.7)$ & $\textcolor{red}{\textbf{93.6}}\,(\pm\,1.9)$ & $\textcolor{red}{\textbf{80.2}}\,(\pm\,3.2)$ & $\textcolor{red}{\textbf{77.3}}\,(\pm\,3.5)$ \\
\midrule
\textbf{GPT}
& AutoSurvey & $90.7\,(\pm\,2.1)$ & $86.2\,(\pm\,2.8)$ & $89.3\,(\pm\,2.5)$ & $84.6\,(\pm\,3.1)$ & $82.3\,(\pm\,3.4)$ \\
& SurveyX & $89.3\,(\pm\,2.6)$ & $84.4\,(\pm\,3.2)$ & $88.5\,(\pm\,2.8)$ & $79.2\,(\pm\,3.9)$ & $78.4\,(\pm\,4.1)$ \\
& SurveyForge & $94.2\,(\pm\,1.8)$ & $87.3\,(\pm\,2.5)$ & $94.8\,(\pm\,1.6)$ & $88.6\,(\pm\,2.7)$ & $88.5\,(\pm\,2.9)$ \\
& SurveyG & $\textcolor{red}{\textbf{95.7}}\,(\pm\,1.5)$ & $\textcolor{red}{\textbf{88.5}}\,(\pm\,2.2)$ & $\textcolor{red}{\textbf{95.1}}\,(\pm\,1.4)$ & $\textcolor{red}{\textbf{92.2}}\,(\pm\,2.3)$ & $\textcolor{red}{\textbf{91.2}}\,(\pm\,2.6)$ \\
\midrule
\textbf{Deepseek}
& AutoSurvey & $86.5\,(\pm\,2.7)$ & $80.4\,(\pm\,3.6)$ & $87.8\,(\pm\,2.9)$ & $77.3\,(\pm\,4.2)$ & $72.4\,(\pm\,4.5)$ \\
& SurveyX & $85.4\,(\pm\,3.1)$ & $81.2\,(\pm\,3.8)$ & $89.4\,(\pm\,2.7)$ & $78.2\,(\pm\,4.3)$ & $75.6\,(\pm\,4.6)$ \\
& SurveyForge & $\textcolor{red}{\textbf{89.4}}\,(\pm\,2.3)$ & $\textcolor{red}{\textbf{86.5}}\,(\pm\,2.9)$ & $92.5\,(\pm\,2.0)$ & $84.1\,(\pm\,3.4)$ & $80.7\,(\pm\,3.7)$ \\
& SurveyG & $88.7\,(\pm\,2.5)$ & $85.7\,(\pm\,3.0)$ & $\textcolor{red}{\textbf{94.2}}\,(\pm\,1.8)$ & $\textcolor{red}{\textbf{86.3}}\,(\pm\,3.1)$ & $\textcolor{red}{\textbf{82.7}}\,(\pm\,3.5)$ \\
\midrule
\textbf{Gemini}
& AutoSurvey & $94.2\,(\pm\,1.9)$ & $70.8\,(\pm\,5.2)$ & $96.5\,(\pm\,1.5)$ & $83.6\,(\pm\,3.6)$ & $84.2\,(\pm\,3.8)$ \\
& SurveyX & $90.2\,(\pm\,2.8)$ & $71.2\,(\pm\,5.4)$ & $95.6\,(\pm\,1.8)$ & $82.8\,(\pm\,3.9)$ & $84.5\,(\pm\,4.0)$ \\
& SurveyForge & $94.9\,(\pm\,1.7)$ & $93.1\,(\pm\,2.1)$ & $98.7\,(\pm\,1.1)$ & $93.5\,(\pm\,2.4)$ & $94.2\,(\pm\,2.5)$ \\
& SurveyG & $\textcolor{red}{\textbf{96.2}}\,(\pm\,1.4)$ & $\textcolor{red}{\textbf{89.4}}\,(\pm\,2.6)$ & $\textcolor{red}{\textbf{98.6}}\,(\pm\,1.0)$ & $\textcolor{red}{\textbf{95.6}}\,(\pm\,2.1)$ & $\textcolor{red}{\textbf{96.2}}\,(\pm\,2.3)$ \\
\bottomrule
\end{tabular}
\caption{LLM-as-a-judge evaluation of generated surveys using GPT-4o-mini backbone (mean (± std) over 10 trials per topic). Each LLM evaluates four generation models across content quality dimensions.}
\label{tab:main_comparison}
\end{table*}
\subsubsection{Full Paper Completion}
In the writing stage, the \textit{Writing Agent} expands each subsection based on the structured outline and its memory, utilizing grounded summaries to ensure factual consistency and contextual relevance. Meanwhile, the \textit{Evaluation Agent} provides critical feedback by offering broader perspectives and generating targeted retrieval queries to identify additional relevant papers from the database $D$. This iterative collaboration ensures that the final text is coherent, comprehensive, and rigorously supported by the literature. The key novelty lies in combining RAG-based retrieval, guided by the Evaluation Agent’s global perspective, with pre-built hierarchical summaries that serve as localized knowledge, enabling the generation of well-balanced and contextually rich subsections.
\section{Experiments}\label{experiment}
\subsection{Experimental Setup}
\subsubsection{Baselines}
We compare \textbf{SurveyG} with three state-of-the-art systems: \textbf{AutoSurvey}~\cite{wang2406autosurvey}, \textbf{SurveyX}~\cite{liang2025surveyx}, and \textbf{SurveyForge}~\cite{yan2025surveyforge}. Details on each baseline are provided in Appendix~\ref{baselines_details}.

\subsubsection{Dataset and Ground Truth Construction} 
We evaluate \textbf{SurveyG} on ten diverse computer science topics from the SurGE benchmark~\cite{su2025benchmarking}, which comprises 205 human-authored surveys and over one million papers. To establish high-quality ground truth, we recruited 20 domain experts: CS Ph.D. students from QS 5-star universities with 3+ years of research experience and senior AI research engineers with publication records in top-tier venues. 

The ground-truth selection followed a rigorous three-stage protocol: (1) \textbf{Topic Selection}: Ten representative research areas covering diverse subfields; (2) \textbf{Survey Selection}: For each topic, 2-3 domain-matched experts independently reviewed available surveys based on four criteria: comprehensive coverage, structural clarity, recency (within 5 years), and citation impact, with disagreements resolved through consensus; (3) \textbf{Reference Curation}: Experts identified seminal works, key methodological contributions, and recent advances essential for comprehensive coverage. The same experts served as human evaluators, assessing coherence, coverage, and factual accuracy. Complete details on topic descriptions, selection criteria, scoring rubrics, and inter-annotator agreement are in Appendix~\ref{ground truth}.

\subsubsection{Implementation Details}
We adopt the experimental setup from prior studies~\cite{yan2025surveyforge, wang2406autosurvey, liang2025surveyx}. Specifically, we retrieve 1,500 candidate papers for outline construction, select 60 papers per chapter, and include 300 papers in each final survey. All methods, including SurveyG and the baselines, use \textit{GPT-4o-mini-2024-07-18} as the main backbone model and share the same crawled paper database released with SurveyForge. To assess cross-model robustness, we additionally test \textit{Gemini-2.5-Flash}. For evaluation, we generate ten independent runs per topic (100 surveys in total). Four state-of-the-art LLMs: \textit{GPT-4o-2024-08-06}, \textit{Claude-3.5-Sonnet-20241022}, \textit{DeepSeek-V3.2-Exp}, and \textit{Gemini-2.5-Pro}, are used as judges. We set $T_{\text{MAX}} = 3$ and DeBERTa-v3-large as our main text encoder model.
 \begin{figure*}[t]  % [t] = top of page (khuyến nghị)
\centering
\includegraphics[width=0.8\textwidth]{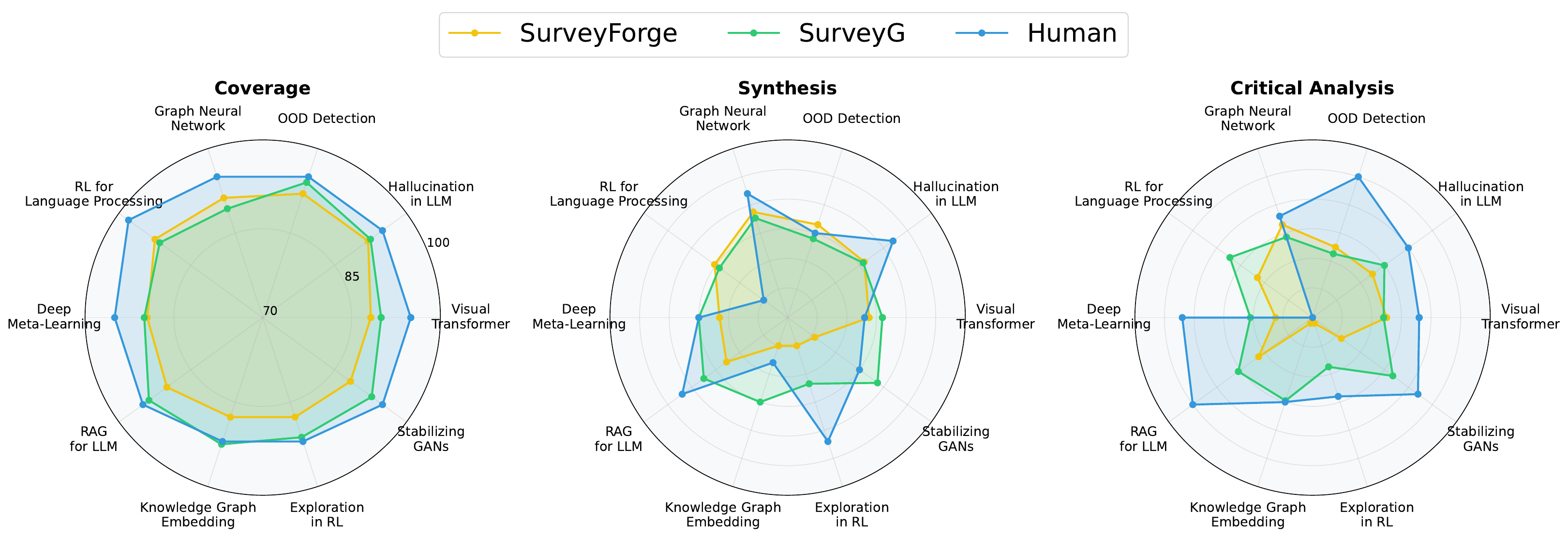}
\caption{LLM-as-a-judge evaluation of human-written ground-truth surveys, SurveyForge, and SurveyG across ten topics using GPT-4o as the evaluator.}
\label{fig:llm_performance}
\end{figure*}
\begin{table*}[ht]
\small
\centering
\begin{tabular}{llccc}
\toprule
\textbf{Evaluation Type} & \textbf{Model} & \textbf{Score Win Rate} & \textbf{Comparative Win Rate} & \textbf{Human Eval} \\
\midrule
\multirow{2}{*}{Full Paper} 
& SurveyForge     & 38.85\% & 27.75\% & 36.00\% \\
& SurveyG (ours)  & \textbf{61.15\%} & \textbf{72.25\%} & \textbf{64.00\%} \\
\midrule
\multirow{2}{*}{Outline} 
& SurveyForge     & 45.00\% & 42.00\% & 45.00\% \\
& SurveyG (ours)  & \textbf{55.00\%} & \textbf{58.00\%} & \textbf{55.00\%} \\
\bottomrule
\end{tabular}
\caption{Win-rate comparison of SurveyForge and SurveyG on full-paper and outline evaluations. \textit{Score Win Rate} indicates higher absolute evaluation scores, \textit{Comparative Win Rate} reflects pairwise LLM preferences, and \textit{Human Eval} denotes human judgment win rate.}
\label{tab:model_comparison}
\end{table*}
\subsubsection{Evaluation Metrics}
We evaluate three dimensions: (1) \textbf{Outline Quality} following~\cite{yan2025surveyforge}; (2) \textbf{Content Quality} using five metrics (\textbf{Coverage}, \textbf{Structure}, \textbf{Relevance}, \textbf{Synthesis}, \textbf{Critical Analysis}) rated 0-100 by LLM and human judges; (3) \textbf{Citation Quality} measuring factual consistency via NLI, reporting \textbf{Citation Recall}, \textbf{Precision}, and \textbf{F1}. Complete metric definitions and evaluation protocols are in Appendix~\ref{evaluation_metrics}.

\subsection{LLMs-as-a-judge evaluation}
\subsubsection{Evaluation on Content Quality}
As shown in Table~\ref{tab:main_comparison}, SurveyG consistently achieves the highest scores across nearly all metrics and evaluation models, demonstrating strong generalization and robustness. SurveyG maintains low standard deviations (1.3-3.7 points) compared to baselines (1.1-5.4 points), indicating stable performance across topics and judges. Cross-judge analysis reveals remarkable consistency: SurveyG achieves top-2 performance under all four judges and ranks first in 16 out of 20 metric-judge combinations (80\%). In contrast, SurveyForge shows variable performance, ranking first under Gemini but dropping to third under Claude's stricter assessment. The particularly great improvements in \textbf{Synthesis} (+2.9 to +4.6 points over SurveyForge) and \textbf{Critical Analysis} (+1.9 to +7.0 points) reflect the framework's ability to integrate information and identify research gaps through structured reasoning, achieving superior organization and analytical depth without requiring human-written survey inputs.
\subsubsection{Evaluation on Ground Truth}
Figure \ref{fig:llm_performance} reveals distinct performance patterns among the three approaches. In Synthesis, SurveyG achieves the most balanced performance, closely matching human surveys and showing more consistent scores than SurveyForge across metrics like OOD Detection and Hallucination in LLM. For Coverage, while human surveys lead with 90 scores, SurveyG demonstrates more stable cross-topic performance compared to SurveyForge's variable results, particularly in specialized areas like Knowledge Graph Embedding and RL for Language Processing. In Critical Analysis, both automated methods score 70-85, but SurveyG shows less variation between metrics, indicating more reliable quality. Overall, while SurveyForge occasionally peaks higher in individual metrics, SurveyG's consistently uniform polygon shapes across all three dimensions suggest superior robustness and generalization capability for diverse survey generation tasks.
\subsubsection{Evaluation on Citation Quality}
As shown in Table \ref{tab:citation}, \textbf{SurveyG} achieves superior citation quality with the highest Recall (91.40$\pm$1.8) and F1 Score (83.49$\pm$2.0), significantly outperforming all baselines. The low standard deviation (1.8-2.0) indicates consistent performance across diverse topics. Notably, SurveyG's Recall approaches the Ground Truth (92.53), demonstrating highly effective identification and linking of relevant literature. While SurveyX achieves the highest Precision (78.12$\pm$2.9), SurveyG's substantially better F1 Score confirms its superior balance between citation completeness and accuracy. This improvement stems from the hierarchical citation graph structure, which enables systematic traversal of foundational and frontier papers, ensuring comprehensive yet relevant reference coverage.

\begin{figure*}[t]  % [t] = top of page (khuyến nghị)
\centering
\includegraphics[width=0.8\textwidth]{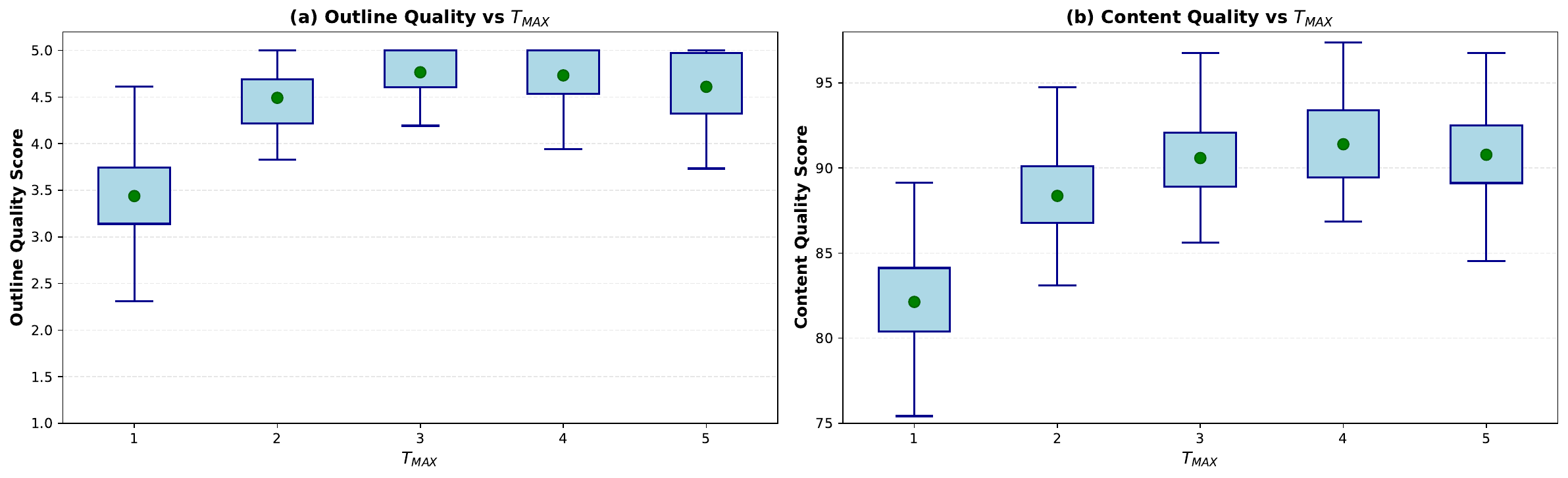}
\caption{Ablation analysis of the number of evaluation loop $T_{MAX}$.}
\label{fig:tmax_ablation}
\end{figure*}

\begin{figure}[t]  % [t] = top of page (khuyến nghị)
\centering
\includegraphics[width=\columnwidth]{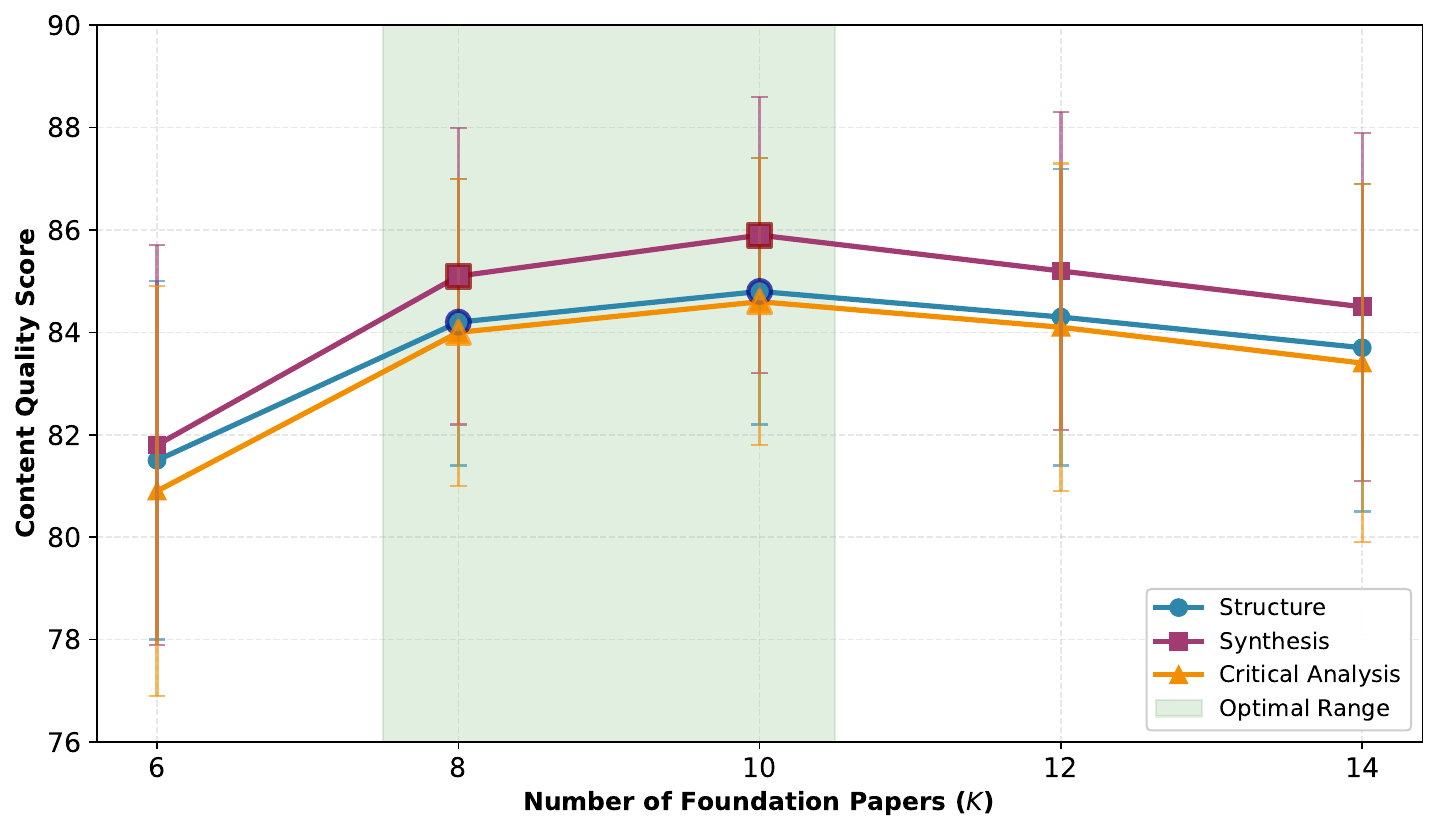}
\caption{Ablation analysis of the number of foundation papers $K$.}
\label{fig:k_ablation}
\end{figure}

\begin{table}[ht]
\centering
\small
\begin{tabular}{lccc}
\toprule
\textbf{Model} & \textbf{Recall} & \textbf{Precision} & \textbf{F1 Score} \\
\midrule
AutoSurvey & $82.25\,(\pm\,2.8)$ & $77.41\,(\pm\,3.2)$ & $79.76\,(\pm\,2.5)$ \\
SurveyForge & $88.34\,(\pm\,2.1)$ & $75.92\,(\pm\,3.5)$ & $81.66\,(\pm\,2.3)$ \\
SurveyX & $85.23\,(\pm\,2.6)$ & \textcolor{red}{$\mathbf{78.12}$}\,$(\pm\,2.9)$ & $81.52\,(\pm\,2.4)$ \\
SurveyG (ours) & \textcolor{red}{$\mathbf{91.40}$}\,$(\pm\,1.8)$ & $77.83\,(\pm\,2.7)$ & \textcolor{red}{$\mathbf{83.49}$}\,$(\pm\,2.0)$ \\
\midrule
Ground Truth & $92.53\,(\pm\,1.5)$ & $86.42\,(\pm\,2.1)$ & $89.34\,(\pm\,1.7)$ \\
\bottomrule
\end{tabular}
\caption{Citation quality comparison across models.}
\label{tab:citation}
\end{table}
\begin{table}[!ht]
\centering
\begin{tabular}{lcc}
\hline
\textbf{Evaluation Pair} & \textbf{Aspect} & Cohen $\kappa$ \\
\hline
LLM vs. Human & Outline & 0.6972 \\
LLM vs. Human & Content & 0.6062 \\
Human Cross-Validation & Outline & 0.7542 \\
Human Cross-Validation & Content & 0.7127 \\
\hline
\end{tabular}
\caption{Inter-rater agreement between LLMs and humans}
\label{tab:agreement}
\end{table}
\subsection{Human Evaluation}
To validate our evaluation framework, we conducted win-rate comparisons between SurveyG and the strongest baseline SurveyForge across ten topics, using three metrics: \textit{Score Win Rate} (higher absolute scores), \textit{Comparative Win Rate} (pairwise preferences), and \textit{Human Eval} (expert judgments). As shown in Table~\ref{tab:model_comparison}, SurveyG substantially outperforms SurveyForge in full-paper generation (61.15\% vs. 38.85\% score win rate, 72.25\% vs. 27.75\% comparative win rate, 64.00\% vs. 36.00\% human preference). The particularly strong comparative win rate (72.25\%) indicates that judges overwhelmingly prefer SurveyG in direct side-by-side comparisons. For outline generation, SurveyG maintains consistency, demonstrating that the hierarchical citation graph provides effective structured guidance from the earliest generation stage. The strong alignment between automated and human judgments (within 3-8\%) validates our LLM-based evaluation framework's reliability while maintaining scalability.
\subsubsection{Details of Human Evaluation}
To evaluate reliability, we assigned two domain experts to each of the ten SurGE topics. For every topic, we randomly sampled ten anonymized outputs from \textbf{SurveyG} and ten from \textbf{SurveyForge}, each obtained from independent generation runs. All outputs were fully anonymized, ensuring that neither the human experts nor the LLM judge (\textit{GPT-4o}) was aware of their system of origin. Both experts independently rated all 20 outputs per topic using identical evaluation prompts and criteria, which covered five content metrics \textbf{Structure}, \textbf{Coverage}, \textbf{Relevance}, \textbf{Synthesis}, and \textbf{Critical Analysis} as well as an outline quality score (0 to 100). For each topic, we computed Cohen’s $\kappa$ to measure (i) agreement between the LLM and human raters, and (ii) agreement between human raters. Table \ref{tab:agreement} reports topic-wise and average $\kappa$ values. The mean Cohen’s $\kappa$ for the outline metric was 0.6972 (LLM-human) versus 0.7542 (human-human), and for content metrics, 0.6062 versus 0.7127, respectively. These results demonstrate substantial inter-rater reliability and confirm that the LLM-as-a-judge evaluations align closely with expert assessments.
\begin{table}[!ht]
\centering
\small
\begin{tabular}{@{}lccccc@{}}
\toprule
% \rowcolor{headercolor!20}
\textbf{Variant} & \textbf{Cov.} & \textbf{Str.} & \textbf{Rel.} & \textbf{Syn.}& \textbf{C.A}\\
\midrule
% \rowcolor{lightgray}
\textbf{Full} & \textbf{95.7} & \textbf{88.5} & \textbf{95.1} & \textbf{92.2} & \textbf{90.5} \\
\midrule
w/o Vertical Traversal & 89.8 & 84.8 & 93.9 & 84.8 & 84.6 \\
% \rowcolor{lightgray!50}
w/o Horizontal Clustering & 91.5 & 85.2 & 93.6 & 83.7 & 86.8 \\
% \rowcolor{lightgray!50}
w/o MA & 93.2 & 85.6 & 91.2 & 88.7 & 87.2 \\
\bottomrule
\end{tabular}
\caption{Ablation study on key components: (1) \textit{w/o Vertical Traversal}: horizontal clustering only; (2) \textit{w/o Horizontal Clustering}: vertical path traversal only; (3) \textit{w/o MA}: no multi-agent refinement. Evaluated by \textit{GPT-4o}.}
\label{tab:ablation-traversal}
\end{table}
\subsection{Ablation Studies}
Table~\ref{tab:ablation-traversal} presents ablation results comparing our full model against variants with specific components removed. The full model achieves the best performance across all metrics, demonstrating that each component contributes meaningfully to survey quality. Removing vertical traversal causes the most significant degradation, particularly in Synthesis (-7.4 points) and Critical Analysis (-5.9 points), indicating that cross-layer exploration from foundational to frontier papers is crucial for integrative reasoning and identifying research gaps. The w/o horizontal clustering variant also shows substantial drops in Synthesis (-8.5 points), confirming that within-layer grouping of thematically related papers enables effective information organization. Removing the multi-agent refinement loop (w/o MA) decreases Relevance (-3.9 points) and Structure (-2.9 points), showing that iterative evaluation-agent feedback improves content focus and coherence. These results validate that the hierarchical citation graph's bidirectional traversal mechanisms and multi-agent refinement work synergistically to produce comprehensive, well-structured surveys.

\textbf{Ablation on number of $T_{\text{MAX}}$}. We investigate the impact of evaluation iterations $T_{\text{MAX}}$ on generation quality (Figure~\ref{fig:tmax_ablation}). For outline generation, performance improves from mean = 3.5 (std = 0.6) at $T_{\text{MAX}} = 1$ to mean = 4.8 (std = 0.30) at $T_{\text{MAX}} = 3$, where iterative feedback effectively addresses structural gaps. Similarly, section content quality increases from 82.5 (std = 3.2) at $T_{\text{MAX}} = 1$ to 90.6 (std = 2.4) at $T_{\text{MAX}} = 3$, achieving full model performance (Structure = 88.5 $\pm$ 2.2, Synthesis = 92.2 $\pm$ 2.3, Critical Analysis = 91.2 $\pm$ 2.6). Beyond $T_{\text{MAX}} = 3$, improvements become marginal (< 0.2 points) with increased variance, indicating diminishing returns. We therefore set $T_{\text{MAX}} = 3$ as default, balancing quality, stability, and computational efficiency.

\textbf{Ablation on number of $K$}. We examine how the number of foundation papers $K$ affects vertical traversal across Structure, Synthesis, and Critical Analysis metrics (Figure~\ref{fig:k_ablation}). Performance is suboptimal at $K = 6$ (scores = 80.9-81.8, std = 3.5-4.0) due to insufficient foundational coverage, improves at $K = 8$ (scores = 84.0-85.1), and peaks at $K = 10$ (scores = 84.6-85.9, std = 2.6-2.8). Synthesis shows the highest sensitivity to $K$ (5.0-point range), as comprehensive foundation-to-frontier paths are critical for cross-paper integration. Optimal $K$ varies by topic breadth: broader fields benefit from $K = 10$, while focused domains achieve comparable results with $K = 8$. Beyond $K = 10$, performance plateaus or declines with increased variance. We recommend $K \in [8, 10]$ based on topic characteristics.

\textbf{Cost Estimation}. The SurveyG framework generates survey papers with an average length of approximately 64k tokens, comparable to expert-written surveys. Each subsection is produced using around 12k input tokens and 800 output tokens. The Evaluation Agent in the RAG loop performs one assessment per subsection, consuming approximately 3.7k input and 700 output tokens. Using GPT-4o-mini pricing (\$0.150/1M input tokens, \$0.600/1M output tokens), the total LLM cost for generating a full 64k-token survey is \$1.5-\$1.7, depending on the number of sections and revision iterations. We provide details about computational efficiency in Appendix \ref{appen:cost}.

\textbf{More ablation studies}. We compare SurveyG and SurveyForge under 2 different backbones: GPT-4o-mini and Gemni-2.5-Flash in Appendix \ref{appen:backbone}. 

\section{Related works}
\subsection{Long-form Text Generation}
LLMs have achieved remarkable progress, yet generating long-form, coherent, and logically structured documents remains a persistent challenge \cite{bai2023longbench, dong2023bamboo, han2024llm}. Recent works have explored different strategies to address the long-context problem. For example, Chain-of-Agents \cite{zhang2024chain} introduces a multi-agent collaboration framework where worker agents process segmented portions of text and a manager agent synthesizes them into coherent outputs, alleviating focus issues in long contexts. LongAlign \cite{bai2401longalign} proposes a recipe for long context alignment, combining instruction data construction, efficient batching, yielding strong gains on queries up to 100k tokens. Complementary to these, Xu et al. \cite{xu2023retrieval} systematically examine the trade-offs between retrieval-augmentation and context-window extension, showing that hybrid approaches can outperform both strategies alone. However, existing approaches often rely on raw reference texts, leading to inefficient retrieval, limited context utilization, and poor structural coherence in survey-like outputs.
\subsection{Automatic Survey Generation}
The automatic generation of literature reviews has been studied for over a decade, starting with multi-document summarization techniques that produced unstructured related work sections. Early systems, such as IBM Science Summarizer \cite{erera2019summarization}, focused on summarizing scientific articles, while more recent LLM-based methods like ChatCite \cite{li2024chatcite} and Susnjak et al. \cite{susnjak2025automating}’s domain-specific fine-tuning advanced the generation of comparative and knowledge-enriched reviews. 
% Xue et al. \cite{xue2026large} propose a large-scale, multidimensional knowledge profiling framework that analyzes over 100,000 ML, vision, and language papers using topic clustering, LLM-assisted parsing to track evolving research themes, methods, and community trends. 
Despite these advances, such methods primarily tackle summarization rather than the creation of fully structured survey papers. More recent systems, including AutoSurvey \cite{wang2406autosurvey}, InteractiveSurvey \cite{wen2025interactivesurvey}, SurveyForge \cite{yan2025surveyforge}, and SurveyX \cite{liang2025surveyx}, propose end-to-end pipelines integrating RAG, clustering, or multi-agent strategies to automate survey construction. These methods improve structural coherence and formatting consistency while scaling to long-form survey content. Nevertheless, most frameworks still restrict users to fixed input-output modes, overlooking relationships among papers and limiting interactivity, which often results in surveys that lack flexibility, relational awareness, and depth.
\section{Limitations and Ethical Considerations}
\paragraph{\textbf{Limitations.}} SurveyG requires substantial computational resources for graph construction and multi-agent refinement, which may limit accessibility for researchers with constrained budgets. The framework's effectiveness depends on the availability and quality of crawled papers, with access limitations to paywalled content potentially creating gaps in coverage. Our evaluation focuses exclusively on English-language computer science publications; generalization to other disciplines or languages remains unexplored.
\paragraph{\textbf{Ethical Considerations.}} Automated survey generation raises concerns about attribution and academic integrity. Users must critically review outputs to ensure proper credit and avoid inadvertent plagiarism. The citation graph inherently reflects existing biases in academic publishing, including systematic underrepresentation of work from marginalized researchers, non-prestigious institutions, and non-English publications.
\section{Conclusion}
In this work, we introduced SurveyG, an automated framework for survey generation that leverages hierarchical knowledge representation and multi-agent collaboration to address the limitations of existing LLM-based approaches. By modeling papers through a three-layer citation-similarity graph and employing both horizontal and vertical traversal strategies, SurveyG captures the structural relationships and evolutionary progress of research, enabling the creation of coherent and well-structured outlines. On the SurGE benchmark for autonomous computer science survey generation, both LLM-as-a-judge evaluations and human expert assessments demonstrate that SurveyG outperforms state-of-the-art frameworks across multiple dimensions.
\newpage
\bibliographystyle{ACM-Reference-Format}
\bibliography{citation}

@article{gao2023enabling,
  title={Enabling large language models to generate text with citations},
  author={Gao, Tianyu and Yen, Howard and Yu, Jiatong and Chen, Danqi},
  journal={arXiv preprint arXiv:2305.14627},
  year={2023}
}

@article{wen2025interactivesurvey,
  title={Interactivesurvey: An llm-based personalized and interactive survey paper generation system},
  author={Wen, Zhiyuan and Cao, Jiannong and Wang, Zian and Guo, Beichen and Yang, Ruosong and Liu, Shuaiqi},
  journal={arXiv preprint arXiv:2504.08762},
  year={2025}
}

@article{liang2025surveyx,
  title={Surveyx: Academic survey automation via large language models},
  author={Liang, Xun and Yang, Jiawei and Wang, Yezhaohui and Tang, Chen and Zheng, Zifan and Song, Shichao and Lin, Zehao and Yang, Yebin and Niu, Simin and Wang, Hanyu and others},
  journal={arXiv preprint arXiv:2502.14776},
  year={2025}
}

@article{wang2406autosurvey,
  title={Autosurvey: Large language models can automatically write surveys, 2024},
  author={Wang, Yidong and Guo, Qi and Yao, Wenjin and Zhang, Hongbo and Zhang, Xin and Wu, Zhen and Zhang, Meishan and Dai, Xinyu and Zhang, Min and Wen, Qingsong and others},
  journal={URL https://arxiv. org/abs/2406.10252}
}

@article{han2024llm,
  title={LLM multi-agent systems: Challenges and open problems},
  author={Han, Shanshan and Zhang, Qifan and Yao, Yuhang and Jin, Weizhao and Xu, Zhaozhuo},
  journal={arXiv preprint arXiv:2402.03578},
  year={2024}
}

@inproceedings{wu2025inference,
  title={Inference scaling laws: An empirical analysis of compute-optimal inference for LLM problem-solving},
  author={Wu, Yangzhen and Sun, Zhiqing and Li, Shanda and Welleck, Sean and Yang, Yiming},
  booktitle={The Thirteenth International Conference on Learning Representations},
  year={2025}
}

@article{su2025benchmarking,
  title={Benchmarking Computer Science Survey Generation},
  author={Su, Weihang and Xie, Anzhe and Ai, Qingyao and Long, Jianming and Mao, Jiaxin and Ye, Ziyi and Liu, Yiqun},
  journal={arXiv preprint arXiv:2508.15658},
  year={2025}
}

@article{huynh2023artificial,
  title={Artificial intelligence for the metaverse: A survey},
  author={Huynh-The, Thien and Pham, Quoc-Viet and Pham, Xuan-Qui and Nguyen, Thanh Thi and Han, Zhu and Kim, Dong-Seong},
  journal={Engineering Applications of Artificial Intelligence},
  volume={117},
  pages={105581},
  year={2023},
  publisher={Elsevier}
}

@article{dong2023bamboo,
  title={Bamboo: A comprehensive benchmark for evaluating long text modeling capacities of large language models},
  author={Dong, Zican and Tang, Tianyi and Li, Junyi and Zhao, Wayne Xin and Wen, Ji-Rong},
  journal={arXiv preprint arXiv:2309.13345},
  year={2023}
}

@article{bai2023longbench,
  title={Longbench: A bilingual, multitask benchmark for long context understanding},
  author={Bai, Yushi and Lv, Xin and Zhang, Jiajie and Lyu, Hongchang and Tang, Jiankai and Huang, Zhidian and Du, Zhengxiao and Liu, Xiao and Zeng, Aohan and Hou, Lei and others},
  journal={arXiv preprint arXiv:2308.14508},
  year={2023}
}

@article{zhang2024chain,
  title={Chain of agents: Large language models collaborating on long-context tasks},
  author={Zhang, Yusen and Sun, Ruoxi and Chen, Yanfei and Pfister, Tomas and Zhang, Rui and Arik, Sercan},
  journal={Advances in Neural Information Processing Systems},
  volume={37},
  pages={132208--132237},
  year={2024}
}

@article{bai2401longalign,
  title={Longalign: A recipe for long context alignment of large language models, 2024},
  author={Bai, Yushi and Lv, Xin and Zhang, Jiajie and He, Yuze and Qi, Ji and Hou, Lei and Tang, Jie and Dong, Yuxiao and Li, Juanzi},
  journal={URL https://arxiv. org/abs/2401.18058}
}

@article{xu2023retrieval,
  title={Retrieval meets long context large language models},
  author={Xu, Peng and Ping, Wei and Wu, Xianchao and McAfee, Lawrence and Zhu, Chen and Liu, Zihan and Subramanian, Sandeep and Bakhturina, Evelina and Shoeybi, Mohammad and Catanzaro, Bryan},
  journal={arXiv preprint arXiv:2310.03025},
  year={2023}
}

@article{erera2019summarization,
  title={A summarization system for scientific documents},
  author={Erera, Shai and Shmueli-Scheuer, Michal and Feigenblat, Guy and Nakash, Ora Peled and Boni, Odellia and Roitman, Haggai and Cohen, Doron and Weiner, Bar and Mass, Yosi and Rivlin, Or and others},
  journal={arXiv preprint arXiv:1908.11152},
  year={2019}
}

@article{li2024chatcite,
  title={ChatCite: LLM agent with human workflow guidance for comparative literature summary},
  author={Li, Yutong and Chen, Lu and Liu, Aiwei and Yu, Kai and Wen, Lijie},
  journal={arXiv preprint arXiv:2403.02574},
  year={2024}
}

@article{susnjak2025automating,
  title={Automating research synthesis with domain-specific large language model fine-tuning},
  author={Susnjak, Teo and Hwang, Peter and Reyes, Napoleon and Barczak, Andre LC and McIntosh, Timothy and Ranathunga, Surangika},
  journal={ACM Transactions on Knowledge Discovery from Data},
  volume={19},
  number={3},
  pages={1--39},
  year={2025},
  publisher={ACM New York, NY}
}

@article{sumers2023cognitive,
  title={Cognitive architectures for language agents},
  author={Sumers, Theodore and Yao, Shunyu and Narasimhan, Karthik and Griffiths, Thomas},
  journal={Transactions on Machine Learning Research},
  year={2023}
}

@inproceedings{wu2024autogen,
  title={Autogen: Enabling next-gen LLM applications via multi-agent conversations},
  author={Wu, Qingyun and Bansal, Gagan and Zhang, Jieyu and Wu, Yiran and Li, Beibin and Zhu, Erkang and Jiang, Li and Zhang, Xiaoyun and Zhang, Shaokun and Liu, Jiale and others},
  booktitle={First Conference on Language Modeling},
  year={2024}
}

@article{yan2025surveyforge,
  title={Surveyforge: On the outline heuristics, memory-driven generation, and multi-dimensional evaluation for automated survey writing},
  author={Yan, Xiangchao and Feng, Shiyang and Yuan, Jiakang and Xia, Renqiu and Wang, Bin and Zhang, Bo and Bai, Lei},
  journal={arXiv preprint arXiv:2503.04629},
  year={2025}
}

@article{traag2019louvain,
  title={From Louvain to Leiden: guaranteeing well-connected communities},
  author={Traag, Vincent A and Waltman, Ludo and Van Eck, Nees Jan},
  journal={Scientific reports},
  volume={9},
  number={1},
  pages={1--12},
  year={2019},
  publisher={Nature Publishing Group}
}

@article{wang2023plan,
  title={Plan-and-solve prompting: Improving zero-shot chain-of-thought reasoning by large language models},
  author={Wang, Lei and Xu, Wanyu and Lan, Yihuai and Hu, Zhiqiang and Lan, Yunshi and Lee, Roy Ka-Wei and Lim, Ee-Peng},
  journal={arXiv preprint arXiv:2305.04091},
  year={2023}
}

@article{zhang2025scientific,
  title={Scientific Paper Retrieval with LLM-Guided Semantic-Based Ranking},
  author={Zhang, Yunyi and Yang, Ruozhen and Jiao, Siqi and Kang, SeongKu and Han, Jiawei},
  journal={arXiv preprint arXiv:2505.21815},
  year={2025}
}
\appendix
\section{Details about Traversal on Graph}\label{traversal}
\begin{algorithm}[t]
\caption{Weighted Breadth-First Search (WBFS)}
\label{alg:wbfs}
\begin{algorithmic}[1]
\State \textbf{Input:} Start node $s$, target layer $\ell$
\State \textbf{Output:} Set of nodes $R$ in layer $\ell$
\State $visited \gets \{s\}$, $queue \gets [s]$, $R \gets \emptyset$
\While{$queue \neq \emptyset$}
    \State $u \gets queue.\textsc{Dequeue}()$
    \ForAll{$v \in \textsc{Successors}(u)$ \textbf{sorted by} $weight(u,v)$ \textbf{desc}}
        \If{$v \notin visited$}
            \State $visited \gets visited \cup \{v\}$
            \If{$v.layer = \ell$}
                \State $R \gets R \cup \{v\}$
            \Else
                \State $queue.\textsc{Enqueue}(v)$
            \EndIf
        \EndIf
    \EndFor
\EndWhile
\State \Return $R$
\end{algorithmic}
\end{algorithm}
We provide a full algorithm of Weighted BFS in Algorithm \ref{alg:wbfs}.
\begin{table*}[t]
\small
\centering
\begin{tabular}{p{0.35\textwidth}p{0.50\textwidth}c}
\hline
\textbf{Topic} & \textbf{Ground Truth Survey} & \textbf{Citation} \\
\hline
Visual Transformer & A Survey of Visual Transformers & 405 \\
\hline
Hallucination in Large Language Models & Siren's Song in the AI Ocean: A Survey on Hallucination in LLMs & 808 \\
\hline
Graph Neural Networks & Graph Neural Networks: Taxonomy, Advances, and Trends & 129 \\
\hline
Deep Meta-Learning & A Survey of Deep Meta-Learning & 459 \\
\hline
Knowledge Graph Embedding & Knowledge graph embedding: A survey from the perspective of representation spaces & 130 \\
\hline
Generalized Out-of-Distribution Detection & Generalized Out-of-Distribution Detection: A Survey & 1406 \\
\hline
Reinforcement Learning for Language Processing & Survey on reinforcement learning for language processing & 206 \\
\hline
Exploration Methods in Reinforcement Learning & Exploration in Deep Reinforcement Learning: From Single-Agent to Multi-Agent Domain & 194 \\
\hline
Stabilizing Generative Adversarial Networks & Stabilizing Generative Adversarial Networks: A Survey & 149 \\
\hline
Retrieval-Augmented Generation for LLMs & Retrieval-Augmented Generation for Large Language Models: A Survey & 953 \\
\hline
\end{tabular}
\caption{Survey Papers Overview}
\label{tab:survey_papers}
\end{table*}
\section{More ablation study results}\label{appen:backbone}
\begin{table*}[ht]
\centering
\small
\begin{tabular}{@{}lllccccc@{}}
\toprule
\textbf{LLM} & \textbf{Model} & \textbf{Backbone} & \textbf{Coverage} & \textbf{Structure} & \textbf{Relevance} & \textbf{Synthesis} & \textbf{Critical Analysis} \\
\midrule
\textbf{Claude}
& SurveyForge & GPT-4o-mini & $81.8\,(\pm\,2.9)$ & $78.4\,(\pm\,3.5)$ & $89.1\,(\pm\,2.1)$ & $75.4\,(\pm\,3.8)$ & $70.3\,(\pm\,4.2)$ \\
& SurveyForge & Gemini-2.5-Flash & $82.5\,(\pm\,3.1)$ & $79.2\,(\pm\,3.7)$ & $89.8\,(\pm\,2.3)$ & $76.8\,(\pm\,3.9)$ & $72.1\,(\pm\,4.3)$ \\
& SurveyG & GPT-4o-mini & $\textcolor{red}{\textbf{88.1}}\,(\pm\,2.4)$ & $87.9\,(\pm\,2.7)$ & $\textcolor{red}{\mathbf{93.6}}\,(\pm\,1.9)$ & $80.2\,(\pm\,3.2)$ & $\textcolor{red}{\textbf{77.3}}\,(\pm\,3.5)$ \\
& SurveyG & Gemini-2.5-Flash & $87.5,(\pm\,2.6)$ & $\mathbf{\textcolor{red}{88.6}\,(\pm\,2.5)}$ & $93,2\,(\pm\,2.0)$ & $\mathbf{\textcolor{red}{81.4}\,(\pm\,3.1)}$ & $76.9\,(\pm\,3.6)$ \\
\midrule
\textbf{GPT}
& SurveyForge & GPT-4o-mini & $94.2\,(\pm\,1.8)$ & $87.3\,(\pm\,2.5)$ & $94.8\,(\pm\,1.6)$ & $88.6\,(\pm\,2.7)$ & $88.5\,(\pm\,2.9)$ \\
& SurveyForge & Gemini-2.5-Flash & $94.8\,(\pm\,1.9)$ & $87.9\,(\pm\,2.6)$ & $95.2\,(\pm\,1.5)$ & $89.4\,(\pm\,2.8)$ & $89.3\,(\pm\,3.0)$ \\
& SurveyG & GPT-4o-mini & $95.7\,(\pm\,1.5)$ & $\textcolor{red}{\mathbf{88.5}}\,(\pm\,2.2)$ & $95.1\,(\pm\,1.4)$ & $\textcolor{red}{\mathbf{92.2}}\,(\pm\,2.3)$ & $\textcolor{red}{\mathbf{91.2}}\,(\pm\,2.6)$ \\
& SurveyG & Gemini-2.5-Flash & $\mathbf{\textcolor{red}{96.3}\,(\pm\,1.4)}$ & $88.2,(\pm\,2.4)$ & $\mathbf{\textcolor{red}{95.7}\,(\pm\,1.3)}$ & $91.8\,(\pm\,2.5)$ & $90.5\,(\pm\,2.5)$ \\
\midrule
\textbf{Deepseek}
& SurveyForge & GPT-4o-mini & $89.4\,(\pm\,2.3)$ & $86.5\,(\pm\,2.9)$ & $92.5\,(\pm\,2.0)$ & $84.1\,(\pm\,3.4)$ & $80.7\,(\pm\,3.7)$ \\
& SurveyForge & Gemini-2.5-Flash & $88.9\,(\pm\,2.5)$ & $85.8\,(\pm\,3.1)$ & $93.1\,(\pm\,2.1)$ & $85.3\,(\pm\,3.5)$ & $81.5\,(\pm\,3.8)$ \\
& SurveyG & GPT-4o-mini & $88.7\,(\pm\,2.5)$ & $85.7\,(\pm\,3.0)$ & $\textcolor{red}{\mathbf{94.2}}\,(\pm\,1.8)$ & $86.3\,(\pm\,3.1)$ & $\textcolor{red}{\mathbf{82.7}}\,(\pm\,3.5)$ \\
& SurveyG & Gemini-2.5-Flash & $\mathbf{\textcolor{red}{89.1}\,(\pm\,2.4)}$ & $\mathbf{\textcolor{red}{86.1}\,(\pm\,2.8)}$ & $93.8\,(\pm\,1.9)$ & $\mathbf{\textcolor{red}{86.8}\,(\pm\,3.0)}$ & $82.3\,(\pm\,3.7)$ \\
\midrule
\textbf{Gemini}
& SurveyForge & GPT-4o-mini & $94.9\,(\pm\,1.7)$ & $93.1\,(\pm\,2.1)$ & $98.7\,(\pm\,1.1)$ & $93.5\,(\pm\,2.4)$ & $94.2\,(\pm\,2.5)$ \\
& SurveyForge & Gemini-2.5-Flash & $95.4\,(\pm\,1.6)$ & $91.8\,(\pm\,2.3)$ & $98.3\,(\pm\,1.2)$ & $94.2\,(\pm\,2.5)$ & $95.1\,(\pm\,2.4)$ \\
& SurveyG & GPT-4o-mini & $96.2\,(\pm\,1.4)$ & $89.4\,(\pm\,2.6)$ & $\textcolor{red}{\mathbf{98.6}}\,(\pm\,1.0)$ & $\textcolor{red}{\mathbf{95.6}}\,(\pm\,2.1)$ & $96.2\,(\pm\,2.3)$ \\
& SurveyG & Gemini-2.5-Flash & $\mathbf{\textcolor{red}{96.5}\,(\pm\,1.3)}$ & $\mathbf{\textcolor{red}{90.1}\,(\pm\,2.3)}$ & $98.4\,(\pm\,1.1)$ & $95.2\,(\pm\,2.3)$ & $\mathbf{\textcolor{red}{96.7}\,(\pm\,2.1)}$ \\
\bottomrule
\end{tabular}
\caption{Backbone comparison: SurveyForge and SurveyG with GPT-4o-mini vs. Gemini-2.5-Flash (mean (± std) over 10 trials per topic). Each LLM evaluates both models with different backbones across content quality dimensions.}
\label{tab:backbone_comparison}
\end{table*}
Table~\ref{tab:backbone_comparison} presents a systematic comparison of backbone models to isolate framework contributions from LLM capability effects. We evaluate both SurveyForge (current SOTA) and SurveyG with two different backbones: GPT-4o-mini and Gemini-2.5-Flash. The results demonstrate that backbone selection has minimal impact on relative performance rankings. For SurveyForge, switching backbones yields marginal differences (average $\pm$0.5-0.8 points across metrics), while SurveyG shows similarly stable performance (average variation $<$0.6 points). Critically, SurveyG maintains consistent superiority over SurveyForge regardless of backbone choice: under Claude's evaluation, SurveyG outperforms SurveyForge by +5.0 to +6.3 points in Coverage and +7.7 to +9.4 points in Structure across both backbones. This pattern holds across all four judges, with SurveyG achieving higher scores in 18 out of 20 model-metric combinations (90\%). The consistency of performance gaps across different backbones confirms that SurveyG's improvements stem from its hierarchical citation graph and multi-level summarization framework rather than simply leveraging more capable foundation models. These findings validate that our architectural innovations, not backbone selection, drive the observed quality improvements in automated survey generation.
\section{Experimental Setting}\label{ground truth}

\subsection{Baseline Systems}\label{baselines_details}
\textbf{AutoSurvey}~\cite{wang2406autosurvey} represents an early multi-stage framework for automated survey generation that combines literature retrieval with structured content synthesis. \textbf{SurveyX}~\cite{liang2025surveyx} enhances AutoSurvey through structured knowledge extraction and outline optimization, introducing refinement mechanisms for improved coherence. \textbf{SurveyForge}~\cite{yan2025surveyforge} leverages human-written survey papers from related domains as prior knowledge for heuristic outline generation, guided by a memory-driven scholar navigation agent that retrieves high-quality references for composing new surveys.

\subsection{Survey Topics and Ground Truth Construction}
We compiled ten representative survey topics covering diverse research areas in machine learning and natural language processing, as summarized in Table~\ref{tab:survey_papers}. Each topic reflects an active line of inquiry, providing a strong foundation for evaluating automated survey generation.

\subsubsection{Expert Recruitment and Qualifications}
Our 20 domain experts comprised:
\begin{itemize}
    \item 12 CS Ph.D. students from QS 5-star universities with 3-5 years of research experience
    \item 8 senior AI research engineers from leading tech companies with 2+ publications in top-tier venues (NeurIPS, ICML, ICLR, ACL, EMNLP, KDD, etc.)
\end{itemize}
Each expert was matched to topics within their domain expertise, ensuring a qualified assessment.

\subsubsection{Three-Stage Ground Truth Selection Protocol}

\textbf{Stage 1: Topic Selection.} We identified ten topics ensuring coverage across diverse AI/ML research areas: (a) \textbf{foundational architectures} (Visual Transformer, Graph Neural Networks, Generative Adversarial Networks), (b) \textbf{learning paradigms} (Deep Meta-Learning, Reinforcement Learning methods), (c) \textbf{knowledge representation and retrieval} (Knowledge Graph Embedding, Retrieval-Augmented Generation), and (d) \textbf{reliability and robustness} (Hallucination in LLMs, Generalized Out-of-Distribution Detection). This selection spans established methodologies, emerging challenges, and cross-domain techniques, ensuring comprehensive evaluation across theoretical foundations, architectural innovations, and practical applications. As shown in Table~\ref{tab:survey_papers}, selected ground-truth surveys are highly cited (mean: 484 citations), indicating strong community recognition.

\textbf{Stage 2: Survey Selection Criteria.} For each topic, 2-3 domain-matched experts independently evaluated candidate surveys from SurGE using four weighted criteria:
\begin{enumerate}
    \item \textbf{Comprehensive Coverage (40\%)}: Survey thoroughly discusses field evolution, seminal works, major methodological branches, and current trends
    \item \textbf{Structural Clarity (25\%)}: Well-organized sections with clear taxonomy, logical flow, and effective use of figures/tables
    \item \textbf{Recency (20\%)}: Published within the last 5 years (2019-2024) to reflect the current research landscape
    \item \textbf{Citation Impact (15\%)}: Highly-cited work (top 20\% within topic area) indicating community recognition
\end{enumerate}
Experts scored each criterion 0-10, with surveys scoring $\geq$ 7.0 average advancing to consensus discussion. Inter-annotator agreement (Fleiss's $\kappa$) averaged 0.73 across topics, indicating substantial agreement.

\textbf{Stage 3: Reference Curation.} Selected experts curated 30-50 essential papers per topic, categorized as: foundational works (seminal papers establishing the field), methodological milestones (key technical innovations), and recent advances (papers from the last 2 years). This reference set serves as ground truth for coverage evaluation.

\subsection{Evaluation Metrics}\label{evaluation_metrics}

\subsubsection{Outline Quality}
Following~\cite{yan2025surveyforge}, we evaluate outline structure, logical coherence, topic coverage, and hierarchical organization on a 0-100 scale. The evaluation prompt assesses whether the outline provides a clear roadmap for comprehensive survey development.

\subsubsection{Content Quality Metrics}
A well-structured outline is essential for maintaining clarity and organization. Full paper evaluation serves as a comprehensive benchmark to assess academic rigor and practical utility. Following~\cite{wang2406autosurvey, liang2025surveyx, yan2025surveyforge, su2025benchmarking}, we assess five metrics on a 0-100 scale:

\textbf{Coverage} measures how thoroughly the survey captures major concepts, foundational works, and emerging trends within the research area.

\textbf{Structure} examines logical organization, section coherence, and taxonomy quality, ensuring readers can navigate the field systematically.

\textbf{Relevance} assesses alignment of content with the target research topic, penalizing off-topic discussions or tangential references.

\textbf{Synthesis} evaluates integration of information from multiple sources into a cohesive, non-redundant narrative that reveals cross-paper insights.

\textbf{Critical Analysis} reflects the survey's ability to identify methodological gaps, highlight comparative trends, and articulate open research challenges.

Each metric is scored independently by both LLM judges and human experts, with final scores averaged across all evaluators.

\subsubsection{Citation Quality Metrics}
Following~\cite{wang2406autosurvey, gao2023enabling}, we measure citation accuracy and contextual relevance. We extract factual claims from generated surveys and verify support using a Natural Language Inference (NLI) model (DeBERTa-v3-large fine-tuned on ANLI).

\textbf{Citation Recall}: Proportion of claims correctly supported by valid references, measuring whether surveys provide adequate citation support.

\textbf{Citation Precision}: Proportion of cited references that truly substantiate their associated claims, measuring citation accuracy.

\textbf{Citation F1}: Harmonic mean of Recall and Precision, providing a balanced measure of citation quality.

For each survey, we sample 100 claim-citation pairs and compute metrics across all samples.
\section{Prompt Templates}\label{prompt}
This section presents the prompt templates designed to guide each stage of automated literature review generation and evaluation. Each template specifies goals, inputs, and evaluation criteria to ensure consistency and quality across generated outputs.

\subsection{Prompt to generate structured outline}
We provide a short version of the prompt template (Figure~\ref{box:prompt_generate_outline}) that instructs the model to construct a coherent, hierarchical outline that captures the logical flow of a literature review topic before detailed writing begins.

\begin{figure}[t]
\begin{tcolorbox}[
    colback=gray!10, 
    colframe=black!40, 
    % fontupper=\small
    % title=Development Path Summarization Prompt,
]
\textbf{Goal}: Generate a structured Literature Review Outline for: "\texttt{[QUERY]}"

\hrule
\vspace{0.5em}

\textbf{INPUT SYNTHESIS DATA}
\begin{itemize}
    \item \textbf{Communities}: \texttt{[PAPER\_COMMUNITIES]}
    \item \textbf{Directions}: \texttt{[DEVELOPMENT\_DIRECTIONS]}
\end{itemize}

\hrule
\vspace{0.5em}

\textbf{REQUIREMENTS \& CONSTRAINTS}

\textbf{1. Structure:}
\begin{itemize}
    \item \textbf{Progression}: Follow Foundations $\to$ Core $\to$ Advanced $\to$ Applications $\to$ Future.
    \item \textbf{Mandatory Sections}: Must include Introduction, Foundational Concepts, and Conclusion.
    \item \textbf{Hierarchy}: Use exactly \textbf{TWO levels} (e.g., 2.1, 2.2). No deeper nesting.
    % \item \textbf{Size}: \texttt{[MIN\_MAIN\_SECTIONS]}-\texttt{[MAX\_MAIN\_SECTIONS]} main sections; \texttt{[MIN\_SUBSECTIONS]}-\texttt{[MAX\_SUBSECTIONS]} subsections each.
\end{itemize}

\textbf{2. Content \& Quality:}
\begin{itemize}
    \item Create a \textbf{coherent narrative} (evolutionary story, not a list).
    \item Group material by \textbf{methodological families} and thematic depth.
    \item Include dedicated sections for Applications and Future Trends/Challenges.
\end{itemize}

\textbf{3. Evidence \& Output:}
\begin{itemize}
    \item \textbf{Proof IDs}: Each subsection \textbf{MUST} be grounded with 1-3 \texttt{proof\_ids} (from \texttt{layer}, \texttt{community\_X}, or \texttt{seed} IDs).
    \item \textbf{Focus Synthesis}: Provide \texttt{section\_focus} (broad theme) and \texttt{subsection\_focus} (specific details).
    \item \textbf{Format}: Return only a \textbf{JSON ARRAY} of main sections and their hierarchical subsections.
\end{itemize}
\end{tcolorbox}
\caption{Generate Outline Prompt.}
\label{box:prompt_generate_outline}
\end{figure}

\subsection{Prompt to evaluate structured outline}
The prompt in Figure \ref{box:prompt_evaluate_outline} guides the model to write complete, citation-based literature review subsections grounded in the provided focus, summaries, and development directions.
The following evaluation prompt extends this process to assess individual sections for depth, synthesis, and analytical quality.
\begin{figure}[t]
\begin{tcolorbox}[
    colback=gray!10, 
    colframe=black!40, 
    % fontupper=\small
    % title=Development Path Summarization Prompt,
]
Evaluate the quality and structure of the following literature review outline. 
Assess whether the outline demonstrates meaningful organization of works rather than a simple concatenation of summaries. \textbf{Your feedback should include:}
\begin{itemize}
    \item Strengths of the outline.
    \item Weaknesses or issues (if any).
    \item Specific suggestions for improvement (only if issues are found).
    \item Final score (1-5, with 5 being the maximum) evaluate overall organization, coherence, and coverage.
\end{itemize}

Outline to evaluate: \{\text{outline\_{text}}\}
\end{tcolorbox}
\caption{Prompt to evaluate structured outline}
\label{box:prompt_evaluate_outline}
\end{figure}

\subsection{Prompt to generate subsections}
This prompt guides the model to write complete, citation-based literature review subsections grounded in the provided focus, summaries, and development directions (Figure~\ref{box:prompt_generate_subsection}).

\begin{figure}[t]
\begin{tcolorbox}[
    colback=gray!10, 
    colframe=black!40, 
    % fontupper=\small
    % title=Development Path Summarization Prompt,
]
\textbf{Task:} Write a comprehensive literature review subsection titled \texttt{[SUBSECTION\_TITLE]} in LaTeX.

\textbf{Inputs:}
\begin{itemize}
    \item \textbf{Focus:} [SUBSECTION\_FOCUS]
    \item \textbf{Community summaries:} [COMMUNITY\_SUMMARY]
    \item \textbf{Development directions:} [DEVELOPMENT\_DIRECTION]
    \item \textbf{Papers (chronological):} [PAPER\_INFO]
\end{itemize}

\textbf{Guidelines:}
\begin{itemize}
    \item Use LaTeX format with citations (\texttt{\textbackslash cite\{citation\_key\}}).
    \item Minimum 400 words, no numbered subsection titles.
    \item Focus strictly on the subsection topic.
    \item Each paper: 2-3 sentences describing technical contributions.
    \item Connect papers by showing how later work addresses earlier limitations.
    \item Conclude with remaining challenges or future directions.
\end{itemize}

\textbf{Avoid:} sequential listing, vague critiques, unsupported claims, isolated descriptions, or ignoring contradictions.

\end{tcolorbox}
\caption{Generate Subsection Prompt}
\label{box:prompt_generate_subsection}
\end{figure}

\subsection{Improve Section Quality}
As shown in Figure~\ref{box:prompt_section_evaluate}, this prompt systematically assesses literature review sections across multiple dimensions, such as content coverage, synthesis, and critical analysis, while offering actionable feedback and retrieval suggestions for refinement.
\begin{figure}[t]
\begin{tcolorbox}[
    colback=gray!10, 
    colframe=black!40, 
    % fontupper=\small
    % title=Development Path Summarization Prompt,
]
Evaluate the quality of the following literature review section within the context of the overall survey outline. 
Your evaluation should address the following aspects, each rated from 1-5 (5 = excellent): 
(1) content coverage, (2) citation density, (3) academic rigor, (4) synthesis across works, 
(5) critical analysis, (6) coherence, (7) depth of discussion, and (8) specificity of scope. 

In addition to numeric ratings, provide:

- A brief natural language summary of the section’s strengths and weaknesses,

- An overall score (1-5) with justification,

- Suggestions for improvement, focusing on areas that fall short,

- A list of search queries that could retrieve additional relevant literature to strengthen the section.

\end{tcolorbox}
\caption{Section Quality and Retrieval Prompt}
\label{box:prompt_section_evaluate}

\end{figure}
\section{Computational Efficiency}\label{appen:cost}
We analyze computational overhead for the hierarchical citation graph operations on a standard workstation (1 NVIDIA RTX 4090, 24GB VRAM): 

\textbf{Paper Retrieval and Graph Construction:} For a typical query retrieving 1,500 candidate papers, the process includes: (1) keyword expansion via LLM (2-3 seconds, \$0.02), (2) Semantic Scholar API calls (15-20 seconds, free), (3) citation network construction (8-12 seconds), and (4) embedding generation using Deberta (45-60 seconds on CPU, negligible GPU cost if available). Total: $\sim$70-95 seconds, \$0.02. 

\textbf{Hierarchical Layer Assignment:} Organizing 300 papers into Foundation/Development/Frontier layers using citation patterns and temporal analysis requires 5-8 seconds (pure algorithmic processing, no API calls). 

\textbf{Multi-Aspect Summarization:} Generating layer-wise and cross-layer summaries involves: (1) horizontal clustering within layers (10-15 seconds for Leiden algorithm), (2) vertical path traversal (3-5 seconds), and (3) LLM-based summarization of $\sim$15-20 paper groups (\$0.80-\$1.20 depending on summary length). Total: $\sim$15-25 seconds computation, \$0.80-\$1.20 LLM cost.

\textit{End-to-End Latency:} The complete pipeline for generating one survey consists of: (1) Crawled data: 20-25 minutes, (2) Graph construction and traversal: 90-128 seconds, (2) Outline generation: 120-180 seconds, (3) Full survey writing: 5-10 minutes (parallel generation of 8-12 sections with evaluation loops). \textbf{Total end-to-end latency: 28-38 minutes per survey}, with the majority of time spent on crawled data rather than graph operations or content generation. 
\section{Case studies}
We provided a subsection generated by \textbf{SurveyG} (Figure~\ref{box:case_study}) to illustrate its ability to synthesize complex research trends in modular and agentic RAG. Overall, this subsection highlights a clear progression in the RAG landscape from simple retrieval pipelines toward multi-stage, agentic, and modular architectures. The discussed works collectively show how LLMs are evolving from passive generators to proactive reasoning agents capable of planning, coordination, and self-optimization. The emergence of meta-frameworks such as AutoRAG and FlashRAG further reflects a shift toward automated orchestration of RAG components, underscoring a broader trend toward unified, adaptive systems that integrate retrieval and reasoning for scalable knowledge synthesis.
\begin{figure*}[t]
\begin{tcolorbox}[
    colback=gray!10, 
    colframe=black!40, 
    % fontupper=\small
    title=Multi-stage and Modular RAG Frameworks,
]

The foundational paradigm of Retrieval-Augmented Generation (RAG) typically operates on a straightforward "retrieve-then-generate" sequence [lewis2020pwr]. However, as Large Language Models (LLMs) are increasingly tasked with complex, multi-faceted queries and dynamic information needs, this simple pipeline proves insufficient [huang2024a59, zhao2024931]. This has spurred the evolution of RAG into more sophisticated, multi-stage, and modular architectures, where the LLM transcends a passive role to become an intelligent agent capable of proactive planning, dynamic decision-making, and the orchestration of various sub-tasks [gao20238ea]. This section focuses on frameworks that empower LLMs to actively manage the information-seeking process through iterative planning, query decomposition, and the dynamic assembly of specialized modules. It is crucial to distinguish these proactive, agentic approaches from reactive or corrective mechanisms (e.g., self-correction, re-ranking) that primarily refine retrieval quality, which are discussed in detail in Section 3.

A significant advancement in modular RAG involves empowering LLMs to act as sophisticated planning agents, iteratively refining their information-seeking process and orchestrating multi-round interactions. [lee2024hif] introduced PlanRAG, which extends the popular ReAct framework by incorporating explicit "Plan" and "Re-plan" steps. This allows LLMs to dynamically generate and iteratively refine analytical approaches based on intermediate retrieval results, effectively acting as decision-makers for complex data analysis tasks. Similarly, [yang20243nb] presented IM-RAG, a multi-round RAG system that leverages learned inner monologues and a multi-agent reinforcement learning approach. In IM-RAG, an LLM-based "Reasoner" dynamically switches between a "Questioner" role (crafting queries) and an "Answerer" role, guided by mid-step rewards from a "Progress Tracker," leading to flexible and interpretable multi-round information gathering. Building on the concept of autonomous interaction, [yu2024c32]'s Auto-RAG enables LLMs to engage in multi-turn dialogues with the retriever, systematically planning retrievals and refining queries until sufficient external information is gathered. This framework highlights the LLM's powerful decision-making capabilities, autonomously adjusting iterations based on query difficulty and knowledge utility. Another approach, [wang2024zt3]'s M-RAG, proposes a multi-partition paradigm for external memories, employing a multi-agent reinforcement learning framework with an "Agent-S" for dynamic partition selection and an "Agent-R" for memory refinement. This enables more fine-grained and focused retrieval by orchestrating memory access across different knowledge partitions. To further optimize the interaction between these modular components, [li20243nz]'s RAG-DDR (Differentiable Data Rewards) offers an end-to-end training method that aligns data preferences between different RAG modules (agents). By collecting rewards and evaluating the impact of perturbations on the entire system, RAG-DDR optimizes agents to produce outputs that enhance overall RAG performance, particularly for smaller LLMs. These agentic frameworks collectively transform RAG into a dynamic, adaptive system capable of tackling complex, multi-hop queries that require sophisticated reasoning and iterative information synthesis.

...

In conclusion, the evolution towards multi-stage and modular RAG frameworks marks a significant advancement, transforming RAG from a simple pipeline into an intelligent, adaptive system. By enabling LLMs to engage in iterative refinement, agentic planning, and dynamic orchestration of sub-tasks, these architectures enhance robustness, reduce hallucinations, and improve the depth and faithfulness of generated responses, particularly for complex, multi-hop queries [tang2024i5r]. However, this sophistication often introduces challenges related to increased computational overhead, the complexity of orchestrating multiple modules, and the need for robust evaluation methodologies that can accurately assess the contributions of each stage and the overall system performance. Benchmarks like [friel20241ct]'s RAGBench, [krishna2024qsh]'s FRAMES, and [tang2024i5r]'s MultiHop-RAG highlight these challenges, emphasizing the need for explainable metrics and unified frameworks to evaluate the intricate interplay of retrieval, reasoning, and generation in these advanced systems. Future research will likely focus on optimizing the efficiency of these multi-stage processes, developing more autonomous and self-correcting agents, and creating more generalized frameworks that can seamlessly integrate diverse knowledge sources and reasoning paradigms while addressing the inherent trade-offs between complexity and efficiency. 

\end{tcolorbox}
\caption{Case studies about the result of the generated subsection.}
\label{box:case_study}
\end{figure*}
\end{document}